\documentclass[runningheads]{llncs}

\usepackage{eccv}

\usepackage{eccvabbrv}

\usepackage{graphicx}
\usepackage{booktabs}

\usepackage[accsupp]{axessibility}  

\usepackage{multirow}
\usepackage[table,xcdraw]{xcolor} 
\usepackage{booktabs}

\usepackage{hyperref}

\usepackage{orcidlink}

\begin{document}

\title{Point Diffusion Mamba: Unified Diffusion-State-Space Modeling for Single-View 3D Reconstruction under Data Scarcity} 

\titlerunning{Point Diffusion Mamba}

\author{Wei Zhou\inst{1}$^\dagger$\orcidlink{0000-0001-8328-3736} \and
Xinzhe Shi\inst{1}$^\dagger$\orcidlink{0009-0008-2356-0541} \and
Xingxing Hao\inst{1}\orcidlink{0000-0002-3590-2392} \and
Xing Hao\inst{1}$^*$\orcidlink{0000-0002-5850-5875} \and
Kang Li\inst{1}$^*$\orcidlink{0000-0001-6218-5715} \and
Jinye Peng\inst{1} \and
Ying He\inst{2}\orcidlink{0000-0002-6749-4485}}

\authorrunning{W. Zhou et al.}

\institute{Northwest University, Xi'an, China \\
\email{zhouwei@nwu.edu.cn}, \email{xinzheshi1@gmail.com} \\ \email{\{xingxing.hao, xhao, likang, pjy\}@nwu.edu.cn},
\and Nanyang Technological University, Singapore\\
\email{yhe@ntu.edu.sg}}

\maketitle

\begingroup
\renewcommand{\thefootnote}{$\dagger$}
\footnotetext{Equal contribution. $^*$ Corresponding author.}
\endgroup

\begin{abstract}
While single-view 3D reconstruction has seen significant pro-\\gress, extrapolating complex 3D structures from inherently ambiguous 2D observations remains fundamentally ill-posed, particularly in the critically underexplored data-scarce regime.
To address this challenge, we propose Point Diffusion Mamba (PDM), a method that integrates the generative power of diffusion models with the efficiency of state-space model for single-view 3D reconstruction under data-scarce conditions. Specifically, PDM employs a lightweight reconstruction module tailored to handle unordered point-cloud inputs effectively. By combining a Local Geometric Aggregation module with Mamba blocks, our approach jointly models global geometric structures and local details. 
In 3D reconstruction, each point in the initial noisy input requires a precise prediction, yet the high-level features extracted by the Mamba module capture only abstract semantic information from sparse points. To bridge this gap, we introduce the Hierarchical Feature Integration Network, which fuses high-level semantic and local geometric features for each point, overcoming the limitations of token-based point-cloud reconstruction. Furthermore, we propose a Dynamic Weighted Sampling strategy that adaptively unifies 3D generation with single-view reconstruction by leveraging generative priors to enhance reconstruction quality. Experimental results on the ShapeNet and Pix3D benchmarks demonstrate that PDM outperforms state-of-the-art methods, providing an effective solution for 3D reconstruction under data-scarce settings.
Code is available at: \href{https://github.com/NWUzhouwei/PDM}{\textcolor{magenta}{\texttt{https://github.com/NWUzhouwei/PDM}}}.
\end{abstract}

\section{Introduction}
\label{sec:intro}

Diffusion-based generative models have recently emerged as a powerful paradigm for 3D modeling, achieving remarkable success in generating detailed and coherent shapes through progressive denoising. These models exhibit unique advantages in challenging tasks such as single-view 3D completion and fine-grained detail reconstruction~\cite{xu2024bayesian,di2023ccd}. For example, PC\textsuperscript{2}~\cite{melas2023pc2} introduces point cloud projection conditioning that maps encoded 2D features back to 3D space during the denoising process, significantly improving single-view reconstruction quality. 

\begin{figure}[h] 
    \centering
    \includegraphics[width=0.62\textwidth]{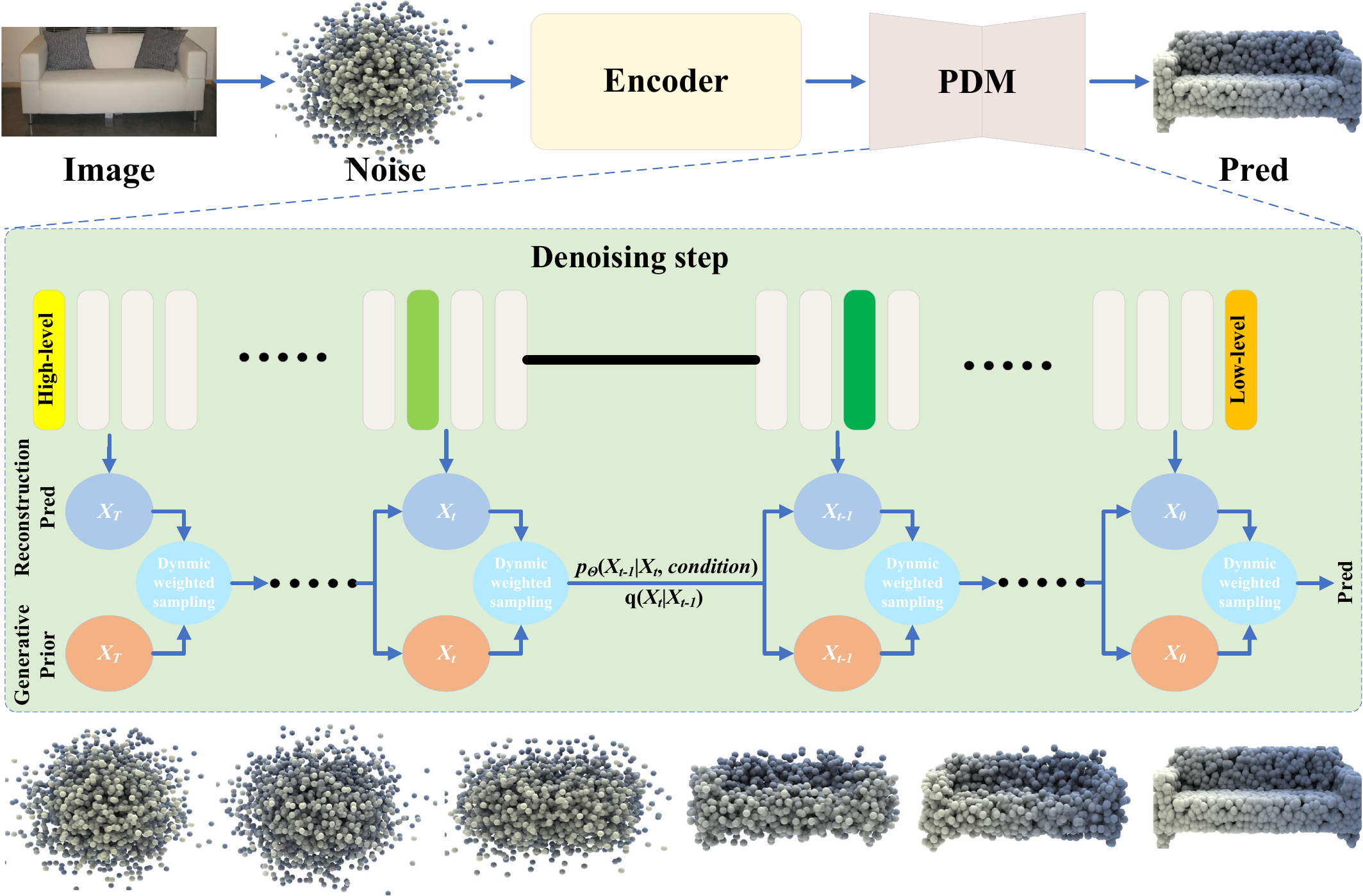}
    \caption{Schematic Illustration of Our PDM. PDM reconstructs original point clouds from noisy inputs. An Encoder extracts image-conditioned features from a single-view image. At each Denoising step, a Dynamic Weighted Sampling (DWS) strategy fuses two representations: the reconstruction prediction and generative prior. Guided by these projected features and DWS, the network learns richer hierarchical geometric priors to progressively generate the final prediction.}
    \label{fig:1}
\end{figure}

\begin{figure}[ht]  
    \centering
    \includegraphics[width=0.6\textwidth]{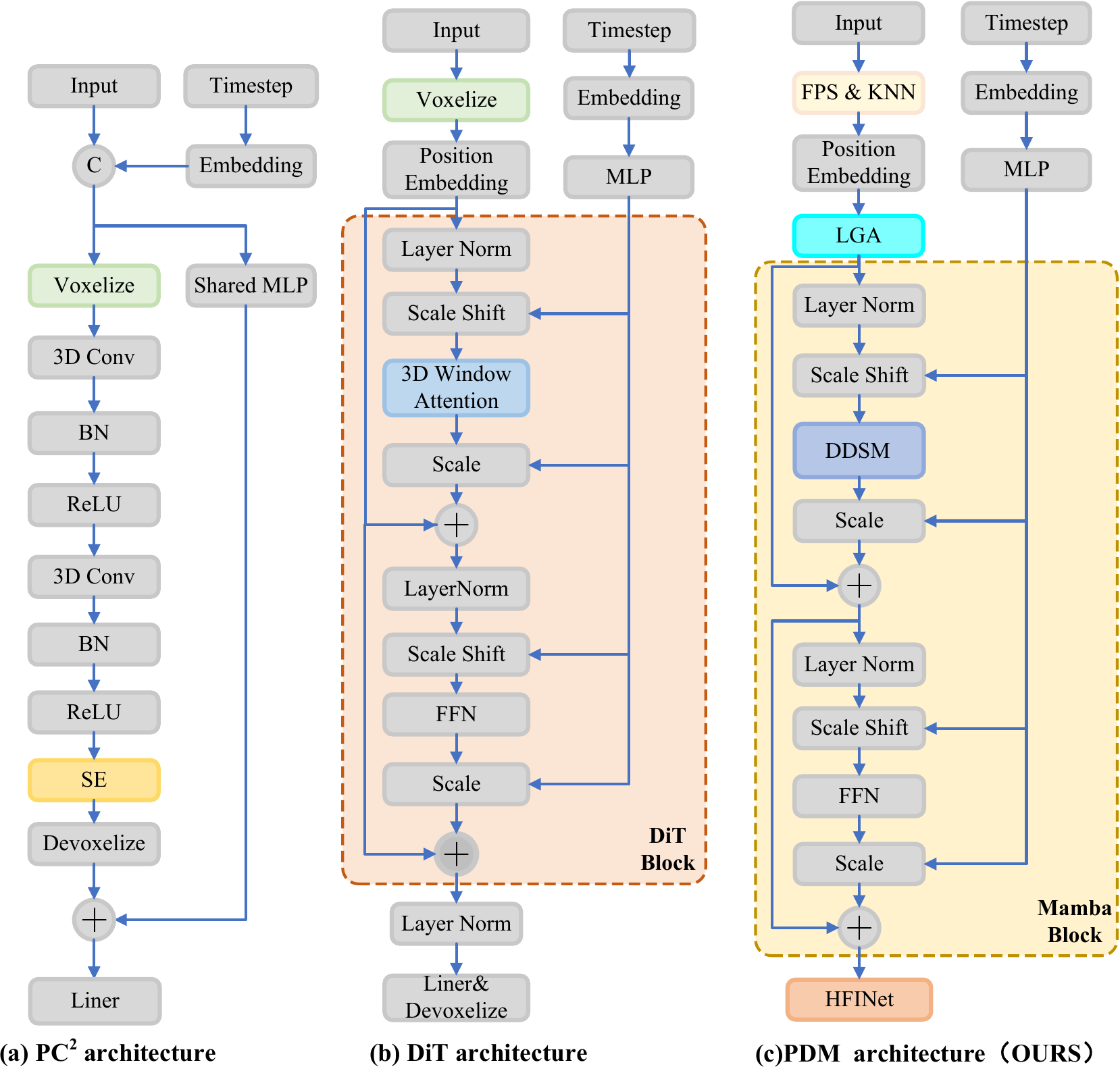}  
    \caption{Different Architectures. (a) The PC\textsuperscript{2} architecture firstly voxelizes the point cloud, then extracts features using 3D convolution, and finally decodes using the channel attention mechanism SE. (b) The DiT architecture also voxelizes the point cloud and organizes it into ordered tokens. It extracts features through 3D window attention and finally decodes them using a linear layer. (c) Our PDM avoids the expensive memory overhead of voxelization by directly using KNN and FPS, while feature extraction is performed using Mamba-diffusion. Finally, HFINet is used for decoding.}
    \label{fig:2}
\end{figure}

Despite these advances, current diffusion-based 3D reconstruction methods still face three key challenges. As shown in Fig.~\ref{fig:2}, MLP-based architectures~\cite{luo2021diffusion,nakayama2023difffacto} struggle to model complex geometric structures, while 3D convolutional architectures~\cite{melas2023pc2,vahdat2022lion,wei2023buildiff} demand substantial memory due to their computational complexity $O(n^3)$ and fail to capture global contextual features. Meanwhile, although Transformer-based architectures~\cite{mo2023dit,mo2024fast} demonstrate promising shape generation capabilities, their attention mechanisms' $O(n^2)$ complexity results in prohibitive memory requirements when processing long sequences.

In 2D high-resolution image generation, state-space models (SSMs) have achieved breakthroughs by reducing computational complexity to a linear scale compared to the quadratic cost of Transformers~\cite{qiao2024hi,ji2025generation,xiao2024frequency,liu2025vmamba}. However, applying SSMs to 3D reconstruction faces these challenges: the unordered nature of point clouds complicates spatial correlations, and SSMs struggle to capture local geometric details. Moreover, while Mamba modules extract high-level semantic features from sparse points, 3D reconstruction demands precise per-point predictions from noisy inputs.

Our proposed framework addresses these limitations through the synergistic integration of diffusion models with state-space model. As depicted in Fig.~\ref{fig:1} and Fig.~\ref{fig:3}, the architecture consists of three functionally complementary components: A lightweight reconstruction module employing Local Geometric Aggregation (LGA) for detail preservation and Mamba blocks for global structure modeling; a Hierarchical Feature Integration Network (HFINet) that enables precise per-point reconstruction by propagating and fusing high-level semantic features; and a Dynamic Weighted Sampling (DWS) strategy that strategically combines generative priors with reconstruction predictions to enhance performance in data-constrained scenarios.

PDM’s core contribution is a task-specific unified framework for single-view 3D reconstruction in the limited-data regime: 
\begin{itemize}
\item Lightweight Mamba-diffusion integration (20.73s runtime, 0.39GB memory) outperforming SOTA (e.g., MESC-3D with 34.05s runtime, 0.86GB memory) while retaining superior geometric accuracy;
\item HFINet for hierarchical feature fusion in diffusion denoising that mitigates inaccurate noise prediction from high-level features alone by integrating them with per-point local features through hierarchical propagation; 
\item DWS with patch-based point correspondence to fuse generative/reconstruction priors, thus to substantially improve reconstruction quality, particularly under data scarcity conditions.
\end{itemize}

\section{Related work}
\label{sec:formatting}

\subsection{\textbf{Single-view Reconstruction}}

Single-view 3D reconstruction aims to recover an object's 3D structure from a single 2D image, a challenging task due to limited depth information. Traditional methods based on projection geometry often fail in complex scenes. Deep learning approaches now dominate, typically employing an encoder-decoder framework with various 3D representations, such as voxel, mesh, point cloud, or implicit functions. However, voxel-based methods suffer from exponential resolution growth~\cite{gao2023cignet,xie2020pix2vox++,tiong20223d}, mesh-based methods struggle with complex shapes~\cite{mao2021std,wen2022pixel2mesh++,yang2023single,zhang2024t}, and implicit methods require extensive multi-view data~\cite{chen2023single,lin2023vision,metzer2023latent}. In contrast, point-based methods balance memory efficiency and detail recovery~\cite{yu2022part,wen20223d,gan20233d,di2023ccd}. SOTA methods like MESC-3D~\cite{li2025mesc} enable point clouds to autonomously select semantic information and incorporates 3D priors through text prompts, and performs well in full-data scenarios.

\subsection{\textbf{Diffusion Models}}

Denoising Diffusion Probabilistic Models (DDPM)~\cite{ho2020denoising} have become a powerful generative modeling framework, excelling in tasks such as image~\cite{graikos2024learned}, text-to-image~\cite{zhu2023conditional}, video~\cite{weng2024art}, and speech generation~\cite{he2024co}. By progressively learning the data distribution, these models mitigate mode collapse and enhance generation diversity. Recently, diffusion models have also been explored for single-view 3D reconstruction~\cite{mu2024gsd,li20243d,liu2024one,long2024wonder3d,zhou2025recurrent}, using conditional generation to infer 3D structures from 2D images. For instance, PC\textsuperscript{2}~\cite{melas2023pc2} uses projection conditioning, projecting local image features onto partially denoised point clouds via rasterization at each diffusion step to ensure geometric consistency.
BDM~\cite{xu2024bayesian} presents a framework coupling top-down prior diffusion and bottom-up data-driven diffusion.
Despite these advancements, diffusion models still face efficiency challenges, particularly due to the quadratic complexity of Transformer-based attention mechanisms.

\subsection{\textbf{State-Space Models}}

SSMs~\cite{hamilton1994state} offer linear-time complexity and efficient long-sequence modeling. The S4 framework~\cite{gu2021efficiently} combines causal convolutions with state-space operations for accurate, scalable sequence analysis, and Mamba~\cite{gu2023mamba} further optimizes memory and computation for superior time-series performance~\cite{xu2024hybrid,behrouz2024graph,wang2024mamba,shams2024ssamba}. These works validate Mamba’s potential in data processing but focus on discriminative analysis tasks, which are orthogonal to our goal of generative single-view 3D reconstruction in the limited-data regime. Our work adapts SSM-based architectures to generative diffusion frameworks, addressing unique challenges of 2D-to-3D lifting that are irrelevant to their discriminative design goals.

\section{Method}
\label{sec:method}

In this section, we introduce the PDM, whose architecture is shown in Fig.~\ref{fig:3}(a). We firstly group a noisy point cloud into different clusters using a grouping module. Then, we use a pre-trained ViT-32~\cite{dosovitskiy2020image} to process the input image and project its features onto the noisy point cloud groups as image conditions. Finally, this condition gradually guides PDM to predict the ground truth point cloud from the noisy input in a point-by-point manner. All the training process and hyperparameter settings of the model are detailed in the code implementation of the supplementary materials.

\subsection{Preliminary: State-Space Model}
\label{sec:ssm}

The State-Space Model (SSM)~\cite{hamilton1994state} is critical for PDM’s efficient sequential data processing. Derived from continuous systems, it maps a 1D sequence \( x(t) \in \mathbb{R} \) to output \( y(t) \in \mathbb{R} \) via a latent state \( h(t) \in \mathbb{R}^N \), governed by matrices \( \textbf{A} \in \mathbb{R}^{N \times N} \), \( \textbf{B} \in \mathbb{R}^{N \times 1} \), and \( \textbf{D} \in \mathbb{R}^{1 \times N} \):
\begin{equation} 
\begin{cases}
    h'(t) = \textbf{A} h(t) + \textbf{B} x(t),\\
    y(t) = \textbf{D} h(t)
\end{cases}
\end{equation}

For discretization, Mamba employs a transformation parameter \( \Delta \) and zero-order hold (ZOH) to convert continuous parameters:
\begin{equation} 
\begin{cases}
\overline{\textbf{A}} = \exp(\Delta \textbf{A}),\\
\overline{\textbf{B}} = (\Delta \textbf{A})^{-1}(\exp(\Delta \textbf{A}) - I) \Delta \textbf{B}
\end{cases}
\end{equation}

The discretized state and output update at each time step is:
\begin{equation} 
\begin{cases}
h_t = \overline{\textbf{A}} h_{t-1} + \overline{\textbf{B}} x_t,\\
y_t = \textbf{D} h_t
\end{cases}
\end{equation}

Notably, the output can be computed via structured convolution, where the kernel \( \overline{\textbf{K}} \) is derived from the state transition matrix:
\begin{equation} 
\begin{cases}
\overline{\textbf{K}} = (\textbf{D}\overline{\textbf{B}}, \textbf{D}\overline{\textbf{A}}\overline{\textbf{B}}, \dots, \textbf{D}\overline{\textbf{A}}^L\overline{\textbf{B}}),\\
y = x * \overline{\textbf{K}}
\end{cases}
\end{equation}
where \( L \) is the length of input \( x \), and \( * \) denotes convolution.

\subsection{Preprocessing}

Owing to the disorder nature of point clouds, they cannot be directly partitioned into contiguous patches like images. To address this inherent disorder, our preprocessing aims to transform unordered points into structured patches. Specifically, we firstly extract the image features using the standard ViT-32 model~\cite{dosovitskiy2020image}. Then these features are projected onto the point cloud through a rasterization function \( P_{\Phi} \), assigning neural features \( Y_{\text{proj}}^t \) to each point based on their spatial positions and camera viewpoints:
\begin{equation} 
Y_{\text{proj}}^t = P_{\Phi}(J, X_t),
\end{equation}
where \( J \) denotes the input image, \( X_t \in \mathbb{R}^{N \times 3} \) represents the noisy point cloud, $N$ is the number of points in the noisy input. 
The projected features are concatenated with the coordinates of \( X_t \), producing a point cloud enhanced with the features $ X_f^t  \in \mathbb{R}^{N \times (3+d_y)}$, and $d_y$ denotes the dimension of the feature.

Similarly to Point-BERT~\cite{yu2022point} and Point-MAE~\cite{pang2022masked}, we partition \( X_f^t \) into patches using a grouping strategy. 
We firstly adopt farthest point sampling (FPS) to obtain center points $\textbf{\textit{C}} \in \mathbb{R}^{s \times 3}$, $s$ is the number of center points. Then we use K-nearest neighbors (KNN) to aggregate \( m \) neighboring points for each center point \( C_i \), thus obtaining patches $\textbf{\textit{P}} \in \mathbb{R}^{s \times m \times (3+d_y)} $:
\begin{equation} 
\begin{cases}
\textbf{\textit{C}}=\{C_i\}_{i=1}^s = \text{FPS}(X_f^t), \\
\textbf{\textit{P}}=\{\textbf{\textit{P}}_i\}_{i=1}^s = \text{KNN}(X_f^t, \textbf{\textit{C}})
\end{cases}
\end{equation}
Finally, we embed the patches $\textbf{\textit{P}}$ in a lightweight encoder \( \xi_{\phi}(\cdot) \) with convolutions and max pooling to obtain patch tokens $\textbf{\textit{F}} \in \mathbb{R}^{s \times d}$:
\begin{equation} 
\textbf{\textit{F}}=\{\textbf{\textit{F}}_i\}_{i=1}^s = \xi_{\phi}(\textbf{\textit{P}})
\end{equation}
where $d$ is the feature dimension.

\begin{figure*}[ht]  
    \centering
    \includegraphics[width=0.9\textwidth]{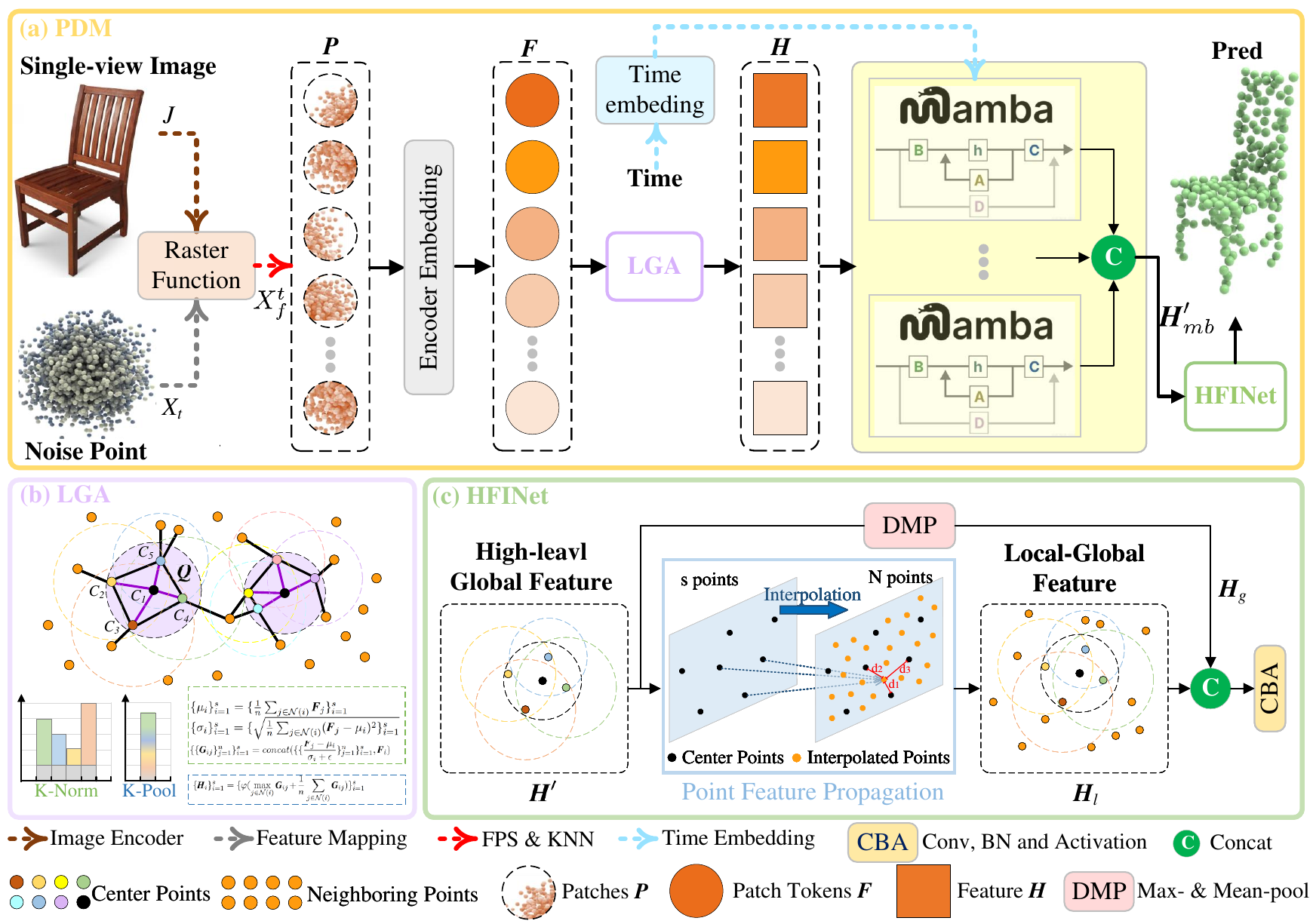}  
    \caption{Our PDM Architecture. (a) The pipeline of our PDM. We firstly extract features from the input image and back-project them onto the point cloud. The point cloud is then divided into point patches and embedded for encoding. Local features of the point cloud are extracted through the LGA module. Next, we learn global features using the Mamba block which encodes temporal conditions through the DiT structure. Finally, we use HFINet to recover the original point cloud from the randomly perturbed point cloud. 
    (b) LGA propagates point features to fuse the features of neighboring points.
    (c) Detailed structure of the HFINet, which fuse high-level features with local features through point feature propagation.
    }
    \label{fig:3}
\end{figure*}

\subsection{Point Diffusion Mamba}
PDM is mainly composed of three components: LGA, Mamba Block, and HFINet.

\paragraph{\textbf{Local geometric aggregation}} The LGA module, as illustrated in Fig.~\ref{fig:3}(b), is proposed to address the limitation of insufficient explicit local geometric feature extraction in Mamba models through feature propagation and neighborhood feature fusion. This module enhances local feature extraction through the coordinated operation of K-Norm
and K-Pool.

K-Norm is performed to establish a neighborhood graph by K-NN for each token $\textbf{\textit{F}}_i$, followed by feature normalization of the neighboring points to ensure consistent feature scaling within the local neighborhood. We firstly adopt K-NN to aggregate $n$ neighboring patch tokens on each token $\textbf{\textit{F}}_i$, thus obtaining patches $\textbf{\textit{Q}} \in \mathbb{R}^{s \times n \times (3+d)}$:
\begin{equation}
    \textbf{\textit{Q}}=\{\textbf{\textit{Q}}_i\}_{i=1}^s = \text{KNN}(\textbf{\textit{F}}, \textbf{\textit{C}})
\end{equation}
Based on $\textbf{\textit{Q}}$, we calculate the mean \( \mu_i \) and standard deviation \( \sigma_i \) for each patch token $\textbf{\textit{F}}_i$ separately:
\begin{equation}
\begin{cases}
    \{\mu_i \}_{i=1}^s=\{ \frac{1}{n} \sum_{j \in \mathcal{N}(i)} \textit{\textbf{F}}_{j}\}_{i=1}^s,\\ 
    \{\sigma_i \}_{i=1}^s=\{ \sqrt{\frac{1}{n} \sum_{j \in \mathcal{N}(i)} (\textit{\textbf{F}}_{j} - \mu_i)^2} \}_{i=1}^s \end{cases}
\end{equation}
where $j$ are the neighboring indices within neighborhood $\mathcal{N}(i)$ of $\textbf{\textit{F}}_i$. Then we normalize the neighboring features and concatenate them with the patch token $\textbf{\textit{F}}_i$, thus obtaining $\{\{\textbf{\textit{G}}_{ij}\}_{j=1}^n\}_{i=1}^s \in \mathbb{R}^{s \times n \times 2d}$: 
\begin{equation}
    \{\{\textbf{\textit{G}}_{ij}\}_{j=1}^n\}_{i=1}^s = \{\{concat(\frac{\textit{\textbf{F}}_{j} - \mu_i}{\sigma_i + \epsilon}, \textbf{\textit{F}}_i)\}_{j=1}^n\}_{i=1}^s
\end{equation}
where \( \epsilon \) is an extremely small constant to prevent division by 0.

After K-Norm, we aggregate the neighboring features using K-Pool. Our K-Pool is a task-specific design for diffusion-based reconstruction,  
which balances salient detail preservation and global smoothness:
\begin{equation}
   \textbf{\textit{H}}=\{\textbf{\textit{H}}_i\}_{i=1}^s = \{\varphi(\max_{j \in \mathcal{N}(i)} \textbf{\textit{G}}_{ij} + \frac{1}{n} \sum_{j \in \mathcal{N}(i)} \textbf{\textit{G}}_{ij})\}_{i=1}^s
\end{equation}
where $\textbf{\textit{H}} \in \mathbb{R}^{s \times d}$, $\varphi$ denotes a simple MLP operation.
This unified aggregation captures both the significant features and the global smoothness , improving the model's ability to represent local geometric information.

\paragraph{\textbf{Mamba block}} 
Unlike images, point clouds exhibit disorder and irregularity, and directly modeling the sequential order may introduce unstable pseudo-sequential dependencies. 
We construct a dual-directional state space module (DDSM) that enhances the model's learning ability from unordered point clouds by integrating both forward sequence modeling $\zeta$ and reverse sequence modeling $\eta$:
\begin{equation}
   \text{DDSM}(\textbf{\textit{H}}) = \textbf{\textit{H}} + \zeta(\textbf{\textit{H}}_c) + \eta(\textbf{\textit{H}}_r) 
\end{equation}
where 
$\textbf{\textit{H}}_c$ is a copy of $\textbf{\textit{H}}$, and $\textbf{\textit{H}}_r$ is the reverse feature obtained by channel flipping.
Unlike typical Mamba, our implementation is redesigned for diffusion modeling: Integrates time-step embeddings to guide noise prediction across diffusion stages; Operates on FPS+KNN grouped patches to preserve local geometric correspondence; Adjusts gating mechanisms to suppress diffusion stochasticity. These modifications ensure DDSM aligns with reconstruction requirements rather than discriminative feature extraction.

To introduce time conditioning to guide the Mamba module in making predictions at different time steps, our module dynamically adjusts the feature distribution of the diffusion model by generating adaptive scaling and shifting parameters through time-conditional adaptive normalization:
\begin{equation}
    \textbf{\textit{H}}_{mb} = \textbf{\textit{H}} + \alpha(t) \cdot \text{DDSM}((\textbf{\textit{H}} \cdot (1 + \beta(t)) + \gamma(t)))
\end{equation}
where $t \in \mathbb{R}^{s \times d}$ is the embedding of the diffusion time step, \( \alpha(t) \) is a gating mechanism that controls the extent of feature updates, \( \beta(t) \) is an adaptive scaling parameter that reflects the expansion or contraction of features, and \( \gamma(t) \) is an adaptive shifting parameter that offsets features.

\paragraph{\textbf{Hierarchical feature integration network}} The high-level features extracted by the Mamba module primarily retain abstract semantic information from sparse points, while 3D point cloud reconstruction requires precise per-point predictions from noisy inputs. 
As shown in Fig.~\ref{fig:3}(c), we propose the HFINet which effectively propagates global high-level features on each point through a feature propagation module, enabling accurate 3D reconstruction via hierarchical local-global feature fusion.

The Mamba block extracts multi-level features $\textbf{\textit{H}}_{mb}'=\{ \textbf{\textit{H}}_{mb_1}, \textbf{\textit{H}}_{mb_2}, ..., \textbf{\textit{H}}_{mb_M} \}$, which are concatenated to form an enhanced feature:
\begin{equation}
\textbf{\textit{H}}' = concat\left(\psi(\textbf{\textit{H}}_{{mb}_1})^{\top}, \psi(\textbf{\textit{H}}_{{mb}_2})^{\top}, ..., \psi(\textbf{\textit{H}}_{{mb}_M})^{\top}\right)
\end{equation}
where $\psi$ denotes the normal layer. Then we adopt max-pooling and mean-pooling to capture global context:
\begin{equation}
\textbf{\textit{H}}_g = concat\left(\text{MaxPool}(\textbf{\textit{H}}'), \text{MeanPool}(\textbf{\textit{H}}')\right)
\end{equation}
Then we use the point feature propagation module $\tau$ to propagate the global feature $\textbf{\textit{H}}'$ throughout the point cloud feature $X_f^t$, thus obtaining local features $H_l$:
\begin{equation}
    \textbf{\textit{H}}_l = \tau(\textbf{\textit{C}}, \textbf{\textit{H}}', X_t, X_f^t)
\end{equation}

The final noise prediction is achieved by fusing the local and global features, which can be represented as:
\begin{equation}
    X_{pred} = \phi(concat(\textbf{\textit{H}}_l, \textbf{\textit{H}}_g))
\end{equation}
where $X_{pred}$ denotes the final noise prediction result, and $\phi$ represents a neural network structure composed of convolutional layers and batch normalization layers and activation functions. This design effectively captures and processes features, enhancing the accuracy and performance of noise prediction.

\subsection{Training Objectives}

In this section, we follow the training strategy of PC \textsuperscript{2}~\cite{melas2023pc2} and define our training objective as:

\begin{equation}
\mathcal{L} = \mathbb{E}_{t, X_0, \epsilon} \left\| \epsilon - \epsilon_\theta \left( \sqrt{\bar{\alpha}_t} X_0 + \\
\sqrt{1 - \bar{\alpha}_t} \epsilon, t \right) \right\|^2
\end{equation}
By minimizing this loss function, we can simultaneously train the PDM. Intuitively, the training process encourages the encoder to extract hierarchical geometric features from the original point cloud and motivates the diffusion model to progressively restore the original point cloud based on these features.

\paragraph{\textbf{Dynamic weighted sampling strategy}} To address the problem of reconstruction quality caused by a limited amount of labeled data, we propose a strategy that improves the quality of point cloud generation by guiding the reconstruction model with the generative model.

As shwon in Fig.~\ref{fig:1}, in the sampling process, instead of applying traditional selection probabilities, we dynamically adjust the fusion weights using a weighted function \( \Phi \) based on the current state of the generative and reconstruction models. This weighting function \( \Phi \) depends on the similarity of the characteristics of each point and the consistency of the output of the two models at the current time step. Specifically, given the point cloud of the generative model \( \mathbf{X}_g \) and the point cloud of the reconstruction model \( \mathbf{X}_r \), we assign a dynamic weight coefficient \( w_j \) to each corresponding point \( \mathbf{x}_{g,j} \) and \( \mathbf{x}_{r,j} \):
\begin{equation}
\label{eq:weight}
w_j = \frac{1}{1 + \exp ( -\delta \cdot \left\| \mathbf{x}_{g,j} - \mathbf{x}_{r,j} \right\|_2^2 )}
\end{equation}
where \( \delta \) is a hyperparameter that controls the sensitivity of the weight, regulating the impact of the difference between the points in the weighting coefficient, in supplementary material, we have conducted the ablation study on \( \delta \) of DWS, 
 \( \mathbf{x}_{g,j} \) and \( \mathbf{x}_{r,j} \) are the coordinates of the \( j \) th point from the generative and reconstruction models.

The final fused point cloud \( \hat{\textbf{X}} \) can then be obtained by weighted averaging as follows:
\begin{equation}
\label{eq:weight_fusion}
\hat{\textbf{X}} = \{ w_j \cdot \textbf{x}_{g,j} + (1 - w_j) \cdot \textbf{x}_{r,j} \}_{j=1}^N
\end{equation}

\begin{figure*}[ht]  
    \centering
    \includegraphics[width=1\textwidth]{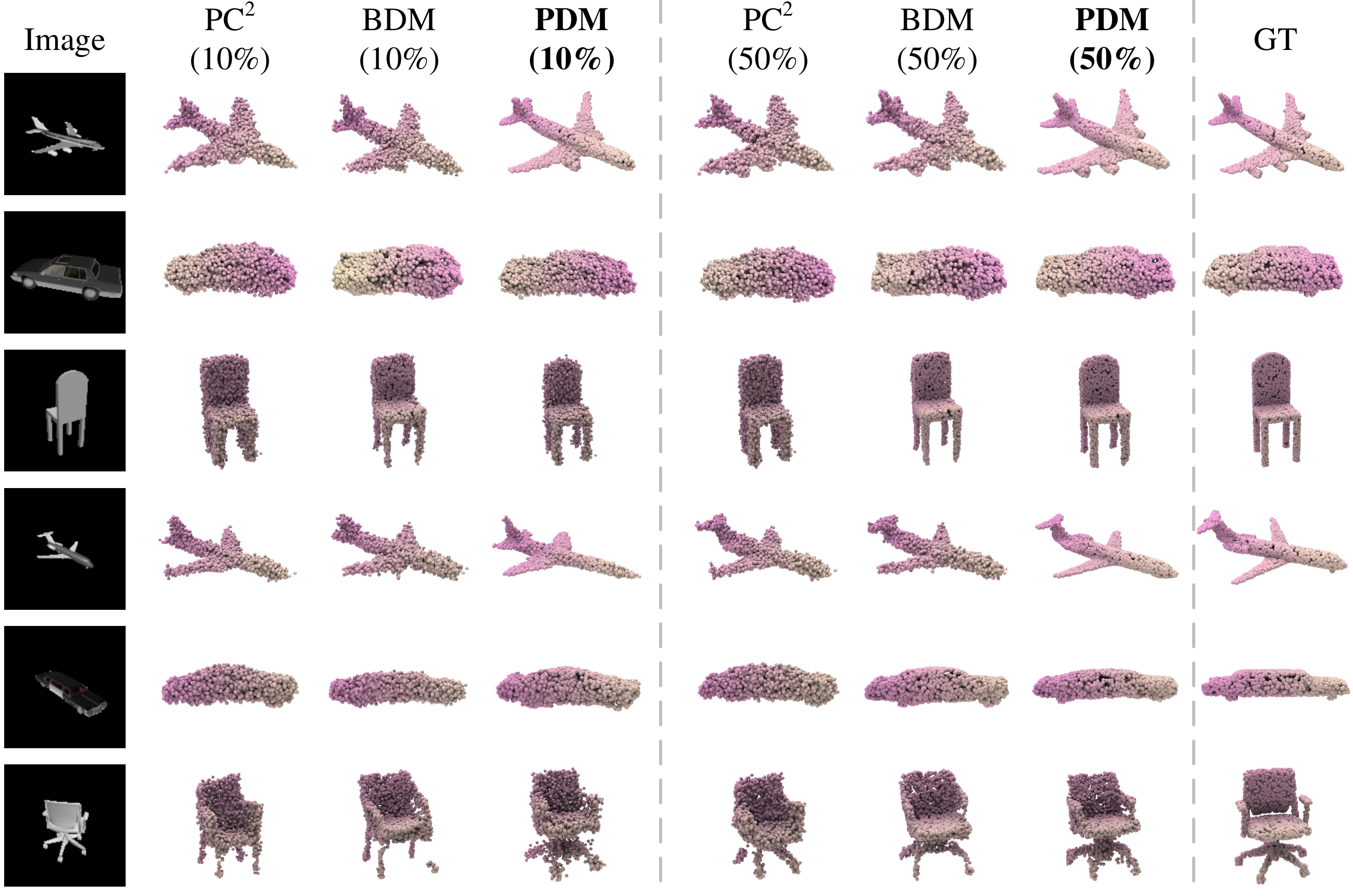}  
    \caption{Qualitative Comparisons on The Synthetic ShapeNet Dataset. Columns 2-4 present the results with 10\% of the data, while columns 5-7 display the results with 50\% of the data. Column 8 provides the corresponding ground truth.}
    \label{fig:shapenet_vis}
\end{figure*}

\section{Experiments}

\subsection{Experimental setup}

\paragraph{\textbf{Experimental datasets}} 
To rigorously evaluate PDM, we replicated the experimental setup introduced in BDM~\cite{xu2024bayesian} by evaluating on both the synthetic the synthetic ShapeNet~\cite{chang2015shapenet} and the real-world Pix3D~\cite{sun2018pix3d}. ShapeNet is a collection of 3D CAD models containing 3,315 categories from the WordNet database. We used a subset of three ShapeNet categories (Chair, Airplane, Car) from 3D-R2N2~\cite{choy20163d} with image renderings, camera matrices, and training/testing splits. 
Likewise, to ensure a fair comparison with SOTA baselines including CCD‑3DR~\cite{di2023ccd}, PC\textsuperscript{2}~\cite{melas2023pc2}, BDM~\cite{xu2024bayesian} and MESC-3D~\cite{li2025mesc}, we evaluated on three Pix3D classes (Chair, Table, Sofa), using their carefully annotated 2D–3D alignments.
We assigned 80\% of the samples for training and reserved the remaining 20\% for testing. 
\footnote{Since CCD‑3DR’s code and the CCD‑backboned BDM variant are not publicly available, we adopted BDM-M / BDM‑B + PC\textsuperscript{2} as our primary baseline.}

\paragraph{\textbf{Implementation details}} For both datasets, we sampled 4096 points for each 3D object and set the rendering resolution to 224$\times$224. Notably, for Pix3D, the images were cropped using their bounding boxes, and the camera matrices were adjusted to accommodate the non-object-centered nature and varying image sizes. For training the generative diffusion model, we used the PDM architecture, removing the image condition to train it as a generative model. 
Generative and reconstruction models are trained independently with no weight sharing and frozen core layers during fusion.
Both the generative and reconstruction models were trained for 140,000 iterations. In our PDM framework, we performed guided point cloud fusion every 32 steps to ensure effective guidance of the generative model during the denoising process. Training of both generative and reconstruction models was conducted on a single NVIDIA GeForce RTX 3090. The train and test logs of MESC-3D and PDM, and the train curves of PDM, PC\textsuperscript{2}, BDM and MESC-3D are provided in the supplementary material.

\subsection{Quantitative Results}

Our evaluation employs two principal metrics for reconstruction quality assessment: Chamfer Distance (CD) and Precision-Recall F-Score (F1@0.01).

\begin{table*}[t]
\centering
\caption{Performance Evaluation on ShapeNet-R2N2. This table compares our PDM with PC\textsuperscript{2}, CCD-3DR, BDM‑M / BDM‑B + PC\textsuperscript{2} and MESC-3D, across three training data scales (10\%, 50\%, and 100\%) using the reconstructed diffusion model. Bold values indicate the best results, while underlined values indicate the second-best results.}
    \renewcommand{\arraystretch}{1.1}
\scalebox{0.57}{
\begin{tabular}{|l|cccccc|cccccc|cccccc|}
\hline
\multicolumn{1}{|c|}{\multirow{3}{*}{method}} & \multicolumn{6}{c|}{Chair} & \multicolumn{6}{c|}{Airplane} & \multicolumn{6}{c|}{Car} \\ \cline{2-19} 
\multicolumn{1}{|c|}{} & \multicolumn{2}{c}{10\%} & \multicolumn{2}{c}{50\%} & \multicolumn{2}{c|}{100\%} & \multicolumn{2}{c}{10\%} & \multicolumn{2}{c}{50\%} & \multicolumn{2}{c|}{100\%} & \multicolumn{2}{c}{10\%} & \multicolumn{2}{c}{50\%} & \multicolumn{2}{c|}{100\%} \\ \cline{2-19} 
\multicolumn{1}{|c|}{} & CD↓ & \multicolumn{1}{c|}{F1↑} & CD↓ & \multicolumn{1}{c|}{F1↑} & CD↓ & F1↑ & CD↓ & \multicolumn{1}{c|}{F1↑} & CD↓ & \multicolumn{1}{c|}{F1↑} & CD↓ & F1↑ & CD↓ & \multicolumn{1}{c|}{F1↑} & CD↓ & \multicolumn{1}{c|}{F1↑} & CD↓ & F1↑ \\ \hline
PC\textsuperscript{2} (CVPR 23)~\cite{melas2023pc2} & 97.25 & \multicolumn{1}{c|}{0.393} & 73.58 & \multicolumn{1}{c|}{0.437} & 65.57 & 0.464 & 88.00 & \multicolumn{1}{c|}{0.605} & 76.39 & \multicolumn{1}{c|}{0.628} & 65.97 & 0.655 & 64.99 & \multicolumn{1}{c|}{0.524} & 62.59 & \multicolumn{1}{c|}{0.542} & 64.36 & 0.547 \\ \cline{2-19} 
CCD-3DR (arXiv 23)~\cite{di2023ccd} & \underline{89.79} & \multicolumn{1}{c|}{\underline{0.418}} & \textbf{63.13} & \multicolumn{1}{c|}{\textbf{0.474}} &\textbf{ 58.47} & \textbf{0.498} & 81.29 & \multicolumn{1}{c|}{\underline{0.612}} & 72.46 & \multicolumn{1}{c|}{0.635} & 62.77 & 0.651 & 63.13 & \multicolumn{1}{c|}{0.531} & 62.25 & \multicolumn{1}{c|}{0.550} & 61.88 & 0.562 \\ \cline{2-19} 
BDM-M (CVPR 24)~\cite{xu2024bayesian} & 94.94 & \multicolumn{1}{c|}{0.395} & 71.56 & \multicolumn{1}{c|}{0.446} & 64.48 & 0.468 & 87.75 & \multicolumn{1}{c|}{0.604} & 73.19 & \multicolumn{1}{c|}{0.629} & 65.16 & 0.653 & 63.53 & \multicolumn{1}{c|}{0.524} & 60.71 & \multicolumn{1}{c|}{0.549} & 64.16 & 0.554 \\ \cline{2-19}
BDM-B (CVPR 24)~\cite{xu2024bayesian} & 94.67 & \multicolumn{1}{c|}{0.410} & 69.99 & \multicolumn{1}{c|}{0.463} & 64.21 & 0.485 & 83.62 & \multicolumn{1}{c|}{\underline{0.612}} & 68.66 & \multicolumn{1}{c|}{\underline{0.641}} & 59.04 & 0.660 & 60.48 & \multicolumn{1}{c|}{\underline{0.539}} & 62.58 & \multicolumn{1}{c|}{\underline{0.554}} & 65.85 & 0.559 \\ \cline{2-19}
MESC-3D (CVPR 25)~\cite{li2025mesc} & 101.51 & \multicolumn{1}{c|}{0.381} & 73.47 & \multicolumn{1}{c|}{0.427} & 65.69 & 0.458 & \underline{74.31} & \multicolumn{1}{c|}{0.611} & \underline{51.28} & \multicolumn{1}{c|}{0.619} & \underline{50.54} & \underline{0.714} & \underline{56.28} & \multicolumn{1}{c|}{0.522} & \textbf{44.48} & \multicolumn{1}{c|}{0.532} & \underline{51.99} & \underline{0.602} \\ \hline
\cellcolor{green!20}PDM & \cellcolor{green!20}\textbf{82.41} & \multicolumn{1}{c|}{\cellcolor{green!20}\textbf{0.419}} & \cellcolor{green!20}\underline{68.85} & \multicolumn{1}{c|}{\cellcolor{green!20}\underline{0.465}}& \cellcolor{green!20}\underline{62.14} & \multicolumn{1}{c|}{\cellcolor{green!20}\underline{0.488}} & \cellcolor{green!20}\textbf{60.64} & \multicolumn{1}{c|}{\cellcolor{green!20}\textbf{0.614}} & \cellcolor{green!20}\textbf{50.14} & \multicolumn{1}{c|}{\cellcolor{green!20}\textbf{0.663}} & \cellcolor{green!20}\textbf{48.66} &\multicolumn{1}{c|}{ \cellcolor{green!20}\textbf{0.719}} & \cellcolor{green!20}\textbf{55.23} & \multicolumn{1}{c|}{\cellcolor{green!20}\textbf{0.542}} & \cellcolor{green!20}\underline{55.53} & \multicolumn{1}{c|}{\cellcolor{green!20}\textbf{0.558}} & \cellcolor{green!20}\textbf{51.54} & \cellcolor{green!20}\textbf{0.607} \\ \hline
\end{tabular}}
\label{quan:shapenet}
\end{table*}

\paragraph{\textbf{ShapeNet}} To directly establish our method's dominance in limited-data environments, we evaluate PDM against state-of-the-art architectures across three dataset scales (10\%, 50\%, 100\%). As detailed in Table~\ref{quan:shapenet}, PDM achieves highly compelling performance, systematically overpowering baselines as data availability decreases. In the critical 10\% low-data regime, PDM's advantages are unmistakable: for Chairs, our CD of 82.41 severely eclipses the second-best CCD-3DR (89.79) by $\sim$9\% and decimates MESC-3D (101.51) by $\sim$23\%, while simultaneously securing the top F1 score (0.419). This dominance extends to Airplanes (CD 60.64 vs. MESC-3D's 74.31) and Cars (CD 55.23 vs. MESC-3D's 56.28). Furthermore, scaling up to 50\% and 100\% data reveals that PDM does not merely memorize sparse datasets but actively scales, maintaining absolute superiority across all Airplane metrics and delivering the best overall performance in the Car 100\% setting (CD 51.54, F1 0.607). These quantitative leaps conclusively validate that PDM provides a vastly more robust prior for single-view reconstruction, especially when data is critically scarce.

\

\begin{table}[t]  
    \centering       
    \caption{Performance Evaluation on Pix3D. This table compares the proposed PDM with PC\textsuperscript{2}, CCD-3DR, BDM-M / BDM‑B + PC\textsuperscript{2} and MESC-3D, using the reconstructed diffusion model and complete training data.}
    \scalebox{0.85}{
    \begin{tabular}{|l|llllll|}
    \hline
    \multirow{2}{*}{Method} & \multicolumn{2}{c|}{Chair} & \multicolumn{2}{c|}{Sofa} & \multicolumn{2}{c|}{Table} \\ \cline{2-7}
     & \multicolumn{1}{c}{CD↓} & \multicolumn{1}{c|}{F1↑} & \multicolumn{1}{c}{CD↓} & \multicolumn{1}{c|}{F1↑} & \multicolumn{1}{c}{CD↓} & \multicolumn{1}{c|}{F1↑} \\ \hline
    PC\textsuperscript{2} (CVPR 23)~\cite{melas2023pc2} & 115.94 & \multicolumn{1}{c|}{0.443} & 47.17 & \multicolumn{1}{c|}{0.445} & 202.77 & 0.397 \\\cline{2-7}
    CCD-3DR (arXiv 23)~\cite{di2023ccd} & 111.42 & \multicolumn{1}{c|}{\underline{0.456}} & 44.91 & \multicolumn{1}{c|}{0.450} & 196.28 & 0.418 \\\cline{2-7}
    BDM-M (CVPR 24)~\cite{xu2024bayesian} & 113.40 & \multicolumn{1}{c|}{0.449} & 44.50 & \multicolumn{1}{c|}{0.451} & 202.08 & 0.413 \\ \cline{2-7}
    BDM-B (CVPR 24)~\cite{xu2024bayesian} & 110.60 & \multicolumn{1}{c|}{0.455} & 45.05 & \multicolumn{1}{c|}{\underline{0.455}} & \underline{186.46} & \textbf{0.429} \\ \cline{2-7}
    MESC-3D (CVPR 25)~\cite{li2025mesc} & \underline{91.36} & \multicolumn{1}{c|}{0.370} & \underline{41.98} & \multicolumn{1}{c|}{0.284} & 206.16 & 0.306 \\ \hline
    \cellcolor{green!20}PDM & {\cellcolor{green!20}\textbf{79.28}} & \multicolumn{1}{c|}{\cellcolor{green!20}\textbf{0.499}} & \cellcolor{green!20}\textbf{41.43} & \multicolumn{1}{c|}{\cellcolor{green!20}\textbf{0.463}} & \cellcolor{green!20}\textbf{184.35} & \cellcolor{green!20}\underline{0.422} \\ \hline
    \end{tabular}}
    \label{quan:pix3d}
\end{table}

\paragraph{\textbf{Pix3D}} Following the exact BDM benchmarking strategy, we evaluate PDM on the real-world Pix3D dataset. As demonstrated in Table~\ref{quan:pix3d}, PDM exhibits remarkable generalization capabilities on in-the-wild captures. For Chairs, PDM registers a CD of 79.28, slashing the error of the next-best method (MESC-3D, 91.36) by an impressive $\sim$15\%, while its F1 score (0.499) eclipses CCD-3DR by $\sim$9.4\%. Similarly, PDM achieves the lowest CD across Sofas (41.43) and Tables (184.35). This comprehensive outperformance strongly suggests that PDM's geometry-aware state-space formulation is uniquely adept at handling complex, non-ideal real-world observations.

In the supplementary material, we provide extensive additional validation: (1) Superior CD and F1 metrics across all remaining ShapeNet categories; (2) A rigorous baseline replacing the PC\textsuperscript{2} Transformer directly with a Mamba block, proving that PDM's massive gains stem from the architectural synergy of LGA, HFINet, and DWS, rather than mere backbone substitution; and (3) Extreme scarcity stress-tests (1\% and 5\% data), where MESC-3D catastrophically collapses while PDM preserves coherent topology.

\subsection{Qualitative Results}

\begin{figure*}[ht]  
    \centering
    \includegraphics[width=1\textwidth]{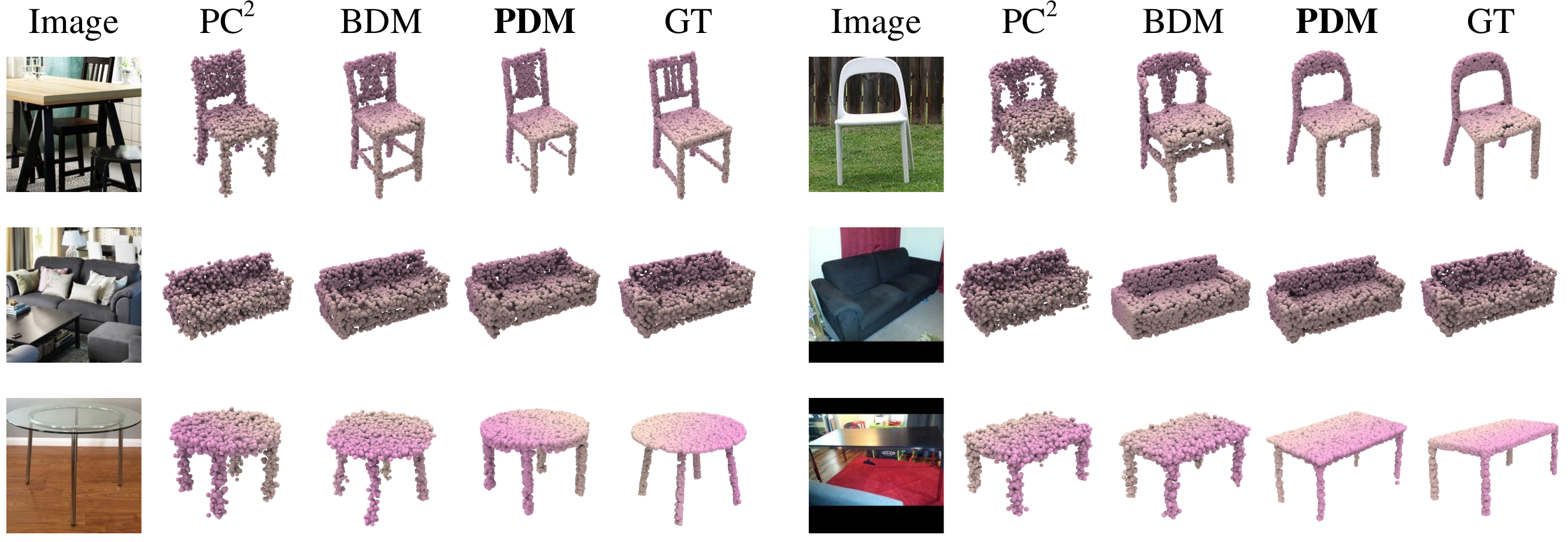}  
    \caption{Qualitative Comparisons on The Real-world Pix3D Dataset under Three Distinct Categories.}
    \label{fig:pix3d_vis}
\end{figure*}

Beyond aggregate metrics, visual inspections starkly highlight PDM's structural superiority. On the ShapeNet dataset (Fig.~\ref{fig:shapenet_vis}), while competing models degenerate into noisy, disjointed clusters under the 10\% data constraint, PDM generates coherent point clouds that strictly preserve global shape semantics and smooth surfaces. Real-world Pix3D visualizations (Fig.~\ref{fig:pix3d_vis}) further cement this advantage: PDM successfully hallucinates and reconstructs fine-grained topological details, such as complex chair backrests, which baseline diffusion models typically oversmooth or fracture entirely.

Moreover, in the supplementary material,  we also demonstrated many other qualitative results on ShapeNet and Pix3D, presented the qualitative results of PDM on unseen real-world captures to show the resilience to occlusion/clutter in the real-world rigid scenarios, evaluated on Objaverse-LVIS to confirm PDM handles diverse geometries effectively, visualized ShapeNet on 1\% and 5\% training samples to verify the performances of PDM on extreme data scarcity, and also visualized all other categories of ShapeNet to demonstrate the effectiveness of our method across all categories.

\subsection{Efficiency}


A core motivation for adopting SSMs is the circumvention of prohibitive attention bottlenecks. As detailed in Table~\ref{table:3}, we benchmarked the parameter count, inference latency, and memory footprint of PDM against leading architectures at a batch size of 1.  Despite a marginal increase in parameter count compared to PC\textsuperscript{2}, PDM achieves a paradigm-shifting reduction in computational overhead. It requires a mere 20.73s runtime and an astonishingly low 0.39GB VRAM, effectively halving the memory footprint of MESC-3D (0.86GB) and running nearly 3$\times$ faster than BDM. This confirms that PDM eliminates the extreme overhead of 3D voxelization and quadratic attention, unlocking high-resolution 3D generation on commodity hardware.

\begin{table}[t] 
    \centering
    \caption{Comparison of Model Efficiency. We compared the model parameters, run time, and GPU memory usage.}
    \scalebox{0.85}{
    \begin{tabular}{|l|c|c|c|} 
        \hline 
        Method & Parameters (M) & Runtime (s) & GPU Memory (GB) \\ \hline 
        PC\textsuperscript{2}/CCD-3DR (CVPR 23) & \textbf{47.41} & 48.46 & 1.73 \\ \cline{2-4}
        BDM-M (CVPR 24) & 74.82 & 52.47 & 2.01 \\ \cline{2-4}
        BDM-B (CVPR 24) & 73.78 & 50.81 & 1.93 \\ \cline{2-4}
        MESC-3D (CVPR 25) & 74.84 & \underline{34.05} & \underline{0.86} \\ \hline
        \cellcolor{green!20}PDM & \cellcolor{green!20}\underline{56.34} & \cellcolor{green!20}\textbf{20.73} & \cellcolor{green!20}\textbf{0.39} \\ \hline 
    \end{tabular}}
    \label{table:3}
\end{table}

\subsection{Ablation Studies}

We systematically dissect the contributions of PDM's architectural innovations. Unless stated otherwise, ablations are executed on the full Pix3D chair subset.


\paragraph{\textbf{Architectural Components}} As reported in the left half of Table~\ref{table:4}, ablating LGA or HFINet induces severe F1 score degradations of 0.021 and 0.064, respectively, empirically validating their indispensable roles in local detail aggregation and global-to-local semantic projection. Crucially, reverting the Mamba blocks to standard self-attention layers drops the F1 by 0.018, confirming Mamba's superior sequence modeling capacity for sparse point clouds. Furthermore, disabling the dual-directional SSM (One-SSM) results in a massive 0.025 F1 penalty, proving that dual-directional state propagation is absolutely critical to overriding the inherent disorder of 3D data.

\paragraph{\textbf{Input patch sequence length}} The right part of Table~\ref{table:4} shows the impact of input patch sequence length on performance. The model performed well at a length of 32, but declined at 64 (likely due to redundant information). Performance improved at 128, with gains tapering off at 256. A further increase to 384 caused another drop, indicating that excessively long sequences hinder information processing. This trend underscores the importance of choosing an appropriate patch sequence length.

\begin{table}[t]
    \centering
    \caption{Ablation Study on Architectural Components and Input Patch Sequence Length. Left: Examines the effects of various modules. Right: Investigates the impact of input sequence lengths (ranging from 32 to 384).}
    \scalebox{0.85}{
    \begin{tabular}{|ccc|ccc|}
        \hline
        Method & F1↑ & CD↓ & Input Size &  F1↑ & CD↓ \\ \hline
        W/O LGA & 0.478 & 91.14 & 32 & 0.485 & 84.31 \\
        W/O HFINet & 0.435 & 126.57 & 64 & 0.474 & 91.15 \\
        Self-Attention & 0.481 & 85.46 & \cellcolor{green!20}128 & \cellcolor{green!20}\textbf{0.499} & \cellcolor{green!20}\textbf{79.28} \\
        One-SSM & 0.474 & 88.72 & 256 & 0.491 & 80.57 \\ 
        \cellcolor{green!20}FULL & \cellcolor{green!20}\textbf{0.499} & \cellcolor{green!20}\textbf{79.28} & 384 & 0.487 & 82.29 \\
        \hline
    \end{tabular}
    }
    \label{table:4}
\end{table}

\paragraph{\textbf{Model scaling analysis}} Left parts of Table~\ref{table:5} reveal scaling patterns across three model variants. PDM-S (9 layers, 192 hidden dim) serves as the baseline. PDM-B (12 layers, 384 hidden dim) improves F1 by 4.2\% and reduces CD by 12.8\%, achieving optimal scaling via increased depth and width. PDM-L (18 layers, 768 hidden dim) shows diminishing returns with slightly lower F1 and CD, indicating that excessive complexity does not guarantee linear performance gains.

\begin{table}[h]
    \centering
    \caption{Ablation Study on Model Size and Sampling Strategy. Left: Evaluates the influence of varying model sizes. Right: Assesses the impact of different sampling strategies.}
    \scalebox{0.85}{
    \begin{tabular}{|ccc|ccc|}
        \hline
        Model & F1↑ & CD↓ & Strategy &  F1↑ & CD↓  \\ \hline
        PDM-S & 0.479 & 90.87 & Direct Sampling & 0.483 & 85.27 \\
        \cellcolor{green!20}PDM-B & \cellcolor{green!20}\textbf{0.499} & \cellcolor{green!20}\textbf{79.28} & BDM Sampling & 0.492 & 80.23 \\
        PDM-L & 0.492 & 80.23 & \cellcolor{green!20}Our Sampling & \cellcolor{green!20}\textbf{0.499} & \cellcolor{green!20}\textbf{79.28} \\
        \hline
    \end{tabular}
    }
     \label{table:5}
\end{table}



\paragraph{\textbf{Dynamic Weighted Sampling Strategy}} The right half of Table~\ref{table:5} validates the necessity of our Dynamic Weighted Sampling (DWS). Direct sampling forms a mediocre baseline (F1: 0.483). While BDM's sampling offers an improvement (F1: 0.492), it fails to intelligently arbitrate between model priors and raw observation. Our DWS explicitly solves this, achieving the definitive best metrics (F1: 0.499, CD: 79.28). By adaptively selecting points based on feature consistency rather than blind spatial distance, DWS actively prevents the degradation of semantically critical regions during the reverse process.

In the supplementary material, we present the ablation for single K-Pool, K-Norm and DWS separately, DWS is employed to adaptively fuse point coordinates based on their spatial consistency and structural agreement between the generative prior and the reconstruction trajectory. This ensures that semantically informative regions are preserved during downsampling. Subsequently, K-Pool aggregates features from local neighbors, allowing the model to capture robust local geometric details and maintain structural consistency across hierarchical levels. 


\section{Conclusions}
\label{sec:conclusion}


Our PDM synergistically integrates diffusion models with efficient of SSM to advance single-view 3D reconstruction, specifically targeting the severely underexplored limited-data regime. By unifying a lightweight module, LGA with dual-directional Mamba blocks, PDM successfully capturing both local details and global structures with sub-quadratic complexity. Furthermore, our proposed HFINet directly resolves the spatial-semantic gap by precisely integrating high-level abstract tokens with localized per-point features. Crucially, our DWS strategy explicitly unifies 3D generation and reconstruction, leveraging hallucinated generative priors to maintain structural integrity. Extensive experiments on the ShapeNet and Pix3D benchmarks demonstrate that PDM significantly outperforms SOTA, establishing a highly robust and memory-efficient paradigm for 3D reconstruction under extreme data scarcity.

\section*{Acknowledgements}
This work was supported in part by the National Key Research and DevelopmentProgram of China under Grant 2024YFF0907604, and in part by the Scientific Research Program of Shaanxi Provincial Education Department under Grant 24JK0674, Natural Science Foundation of Shaanxi Province under Grant 2025JC-YBQN-889 and National Natural Science Foundation of China under Grant 62572394.

%
%
\bibliographystyle{splncs04}
\bibliography{main}
\end{document}